\documentclass[letterpaper, 10 pt, conference]{ieeeconf}  % Comment this line out if you need a4paper
\usepackage{xifthen}
\usepackage{xparse}
\usepackage{subdepth}
\usepackage{leftidx}
\usepackage{bm}
\newcommand{\bbm}{\begin{bmatrix}}
\newcommand{\ebm}{\end{bmatrix}}

\DeclareMathAlphabet{\mbf}{OT1}{ptm}{b}{n}
\newcommand{\mbs}[1]{{\bm{#1}}} 

\newcommand{\mbsbar}[1]{{\overline{\boldsymbol{#1}}}}
\newcommand{\mbshat}[1]{{\hat{\boldsymbol{#1}}}}
\newcommand{\mbstilde}[1]{{\tilde{\boldsymbol{#1}}}}

\newcommand{\mbsdot}[1]{{\dot {\boldsymbol{#1}}}}

\newcommand{\mbfbar}[1]{{\overline{\mbf{#1}}}}
\newcommand{\mbfhat}[1]{{\hat{\mbf{#1}}}}
\newcommand{\mbftilde}[1]{{\tilde{\mbf{#1}}}}

\newcommand{\mbfdot}[1]{{\dot{\mbf{#1}}}}

\newcommand{\cframe}[1]{{\smash{\protect\underrightarrow{\mathcal{F}}_{#1}}}}

\DeclareMathAlphabet{\mathbfit}{OML}{cmm}{b}{it}
\newcommand{\homo}[1]{{\mathbfit{#1}}}

\newcommand{\mbfh}[1]{{\homo{#1}}}

\newcommand{\vel}[3]{\leftidx{_{#1}}{\mbf v}{\IfValueTF{#2}{_{#2#3\hspace{2pt}}}{}}} % velocity
\newcommand{\veltilde}[3]{\leftidx{_{#1}}{\mbftilde v}{\IfValueTF{#2}{_{#2#3\hspace{2pt}}}{}}} % velocity
\newcommand{\velbar}[3]{\leftidx{_{#1}}{\mbfbar v}{\IfValueTF{#2}{_{#2#3\hspace{2pt}}}{}}} % velocity
\newcommand{\velhat}[3]{\leftidx{_{#1}}{\mbfhat v}{\IfValueTF{#2}{_{#2#3\hspace{2pt}}}{}}} % velocity
\newcommand{\veldot}[3]{\leftidx{_{#1}}{\mbfdot v}{\IfValueTF{#2}{_{#2#3\hspace{2pt}}}{}}} % velocity

\newcommand{\acc}[3]{\leftidx{_{#1}}{\mbf a}{\IfValueTF{#2}{_{#2#3\hspace{2pt}}}{}}} % acceleration
\newcommand{\acctilde}[3]{\leftidx{_{#1}}{\mbftilde a}{\IfValueTF{#2}{_{#2#3\hspace{2pt}}}{}}} % acceleration
\newcommand{\accbar}[3]{\leftidx{_{#1}}{\mbfbar a}{\IfValueTF{#2}{_{#2#3\hspace{2pt}}}{}}} % acceleration

\newcommand{\rotvel}[3]{\leftidx{_{#1}}{\mbs \omega}{\IfValueTF{#2}{_{#2#3\hspace{2pt}}}{}}} % rotational velocity
\newcommand{\rotveltilde}[3]{\leftidx{_{#1}}{\mbstilde \omega}{\IfValueTF{#2}{_{#2#3\hspace{2pt}}}{}}} % rotational velocity
\newcommand{\rotvelbar}[3]{\leftidx{_{#1}}{\mbsbar \omega}{\IfValueTF{#2}{_{#2#3\hspace{2pt}}}{}}} % rotational velocity
\newcommand{\rotvelhat}[3]{\leftidx{_{#1}}{\mbshat \omega}{\IfValueTF{#2}{_{#2#3\hspace{2pt}}}{}}} % rotational velocity
\newcommand{\rotveldot}[3]{\leftidx{_{#1}}{\mbsdot \omega}{\IfValueTF{#2}{_{#2#3\hspace{2pt}}}{}}} % rotational velocity derivative

\newcommand{\T}[2]{\leftidx{}{\mbfh T}{_{#1#2\hspace{2pt}}}} % homogeneous transformation matrix
\IEEEoverridecommandlockouts                              % This command is only needed if 
\usepackage{hyperref}

\title{\LARGE \bf
REGRACE: A Robust and Efficient Graph-based Re-localization Algorithm using Consistency Evaluation
}

\author{Débora N.P. Oliveira$^{1,\ast}$, Joshua Knights$^{2,\ast}$, Sebastián Barbas Laina$^{3,4,\ast}$, Simon Boche$^{3,4,\ast}$, \\Wolfram Burgard$^{1}$ and Stefan Leutenegger$^{3,4,5}$% <-this % stops a space
\thanks{${1}$ Artificial Intelligence and Robotics Lab, University of Technology of Nuremberg (UTN). \{\tt\small\textit{firstname.lastname}\}\tt\small{@utn.de}}%
\thanks{${2}$ CSIRO Robotics, DATA61, and Queensland University of Technology (QUT). \{\tt\small\textit{firstname.lastname}\}\tt\small{@csiro.au}}%
\thanks{${3}$ Mobile Robotics Lab, Technical University of Munich (TUM). \{\tt\small\textit{firstname.lastname}\}\tt\small{@tum.de}}%
\thanks{${4}$ Munich Institute of Robotics and Machine Intelligence (MIRMI).}%
\thanks{${5}$ Mobile Robotics Lab, ETH Zurich.}%
\thanks{${\ast}$ Contributed equally to the work as first co-author.}%
\thanks{${\dagger}$ Code and trained weights available at \url{https://github.com/ethz-mrl/regrace}.}%
}

\usepackage{lipsum}  
\usepackage{multirow}
\usepackage{amsmath}
\usepackage{amsfonts}
\usepackage[table,xcdraw]{xcolor}
\usepackage{arydshln}
\usepackage{booktabs}
\usepackage{tikz}\usepackage[caption=false, font=footnotesize]{subfig}

\begin{document}

\maketitle
\thispagestyle{empty}
\pagestyle{empty}

%%%%%%%%%%%%%%%%%%%%%%%%%%%%%%%%%%%%%%%%%%%%%%%%%%%%%%%%%%%%%%%%%%%%%%%%%%%%%%%%
{\begin{abstract}

Loop closures are essential for correcting odometry drift and creating consistent maps, especially in the context of large-scale navigation. Current methods using dense point clouds for accurate place recognition do not scale well due to computationally expensive scan-to-scan comparisons. Alternative object-centric approaches are more efficient but often struggle with sensitivity to viewpoint variation. In this work, we introduce REGRACE,  a novel approach that addresses these challenges of scalability and perspective difference in re-localization by using LiDAR-based submaps. We introduce rotation-invariant features for each labeled object and enhance them with neighborhood context through a graph neural network. To identify potential revisits, we employ a scalable bag-of-words approach, pooling one learned global feature per submap. Additionally, we define a revisit with geometrical consistency cues rather than embedding distance, allowing us to recognize far-away loop closures. Our evaluations demonstrate that REGRACE achieves similar results compared to state-of-the-art place recognition and registration baselines while being twice as fast. Code and models are publicly available$^\dagger$.

\end{abstract}}

\bstctlcite{IEEEexample:BSTcontrol}
%%%%%%%%%%%%%%%%%%%%%%%%%%%%%%%%%%%%%%%%%%%%%%%%%%%%%%%%%%%%%%%%%%%%%%%%%%%%%%%%
\section{INTRODUCTION}

In the re-localization problem, a mobile robot living in the three-dimensional world needs to estimate a six degrees of freedom (DoF) transformation to align the current trajectory with past routes. In practice, this registration is typically done by comparing keypoints and features between the current sensor input and previously seen images or LiDAR scans~\cite{Gutmann1998experimental}. However, doing this based on individual observations suffers from perspective problems, as keypoint descriptors must remain similar despite occlusions and viewpoint differences.

To provide a more geometrically accurate and lightweight representation, research works have used submaps in SLAM~\cite{boche2024tightly} and re-localization~\cite{yuan2024btc,Cramariuc2021SemSegMap3}. These dense partial maps are created by integrating LiDAR scans from multiple viewpoints into the same frame, which reduces the dependency on perspective. Nonetheless, obtaining a compact yet informative representation of submaps remains challenging. 

Object-centric graph-based re-localization~\cite{Pramatarov2022BoxGraphSP,kong2020sgpr, wang2024sglc} has demonstrated scalability when extracting descriptors, but graph-to-graph comparison remains a bottleneck. This matching is often achieved using the Hungarian algorithm~\cite{Pramatarov2022BoxGraphSP}, which experiences longer runtimes as the number of nodes increases. Alternative works reduce runtime by learning a similarity score between graphs~\cite{kong2020sgpr,wang2024sglc}, but still employ perspective-dependent features or handcrafted descriptors that struggle with domain change.  Moreover, most research has focused on short-range \cite{vid2022logg3d,kim2021scan} instead of long-range loop closure, which is particularly required in unstructured environments where staying near previously traveled paths can not be guaranteed.

\begin{figure}[t]
      \centering 
      \includegraphics[width=.95\linewidth]{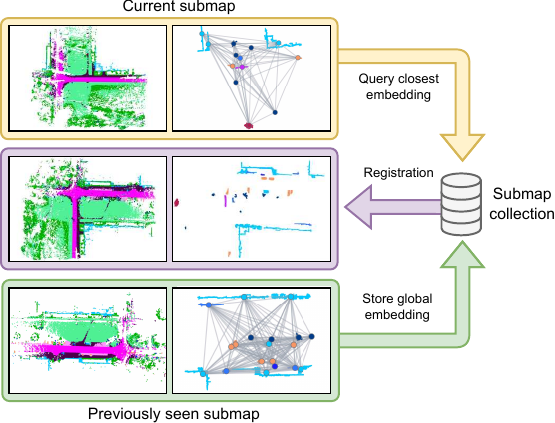}
      \caption{Example of REGRACE on KITTI sequence 08. We use a bag-of-words query over global embeddings pooled from a graph representation.} 
      \label{fig:kitti-08} 
\end{figure}

This paper proposes REGRACE, a scalable place recognition and registration pipeline for submaps. We produce dense submaps by aggregating LiDAR scans into a single frame, expanding the sensor field of view and reducing object occlusion. We also propose context-aware object descriptors by representing each submap as a graph , and introduce a new approach to detecting loop closures, which uses geometric consistency to classify revisits rather than embedding distance, helping detect far-away revisits. Fig.~\ref{fig:kitti-08} shows a typical result obtained with REGRACE. In summary:
\begin{itemize}
    %\item We propose REGRACE, a scalable, learning-based, object-centric re-localization pipeline for submaps.
    \item We introduce context-aware global descriptors extracted from rotation-invariant object features using a lightweight graph neural network.
    \item We propose identifying revisits based on geometric consistency instead of embedding space distance, enhancing performance across all evaluated methods.
    \item Our proposed graph-based method produces fewer local keypoints than the baselines, significantly reducing the time required for registration.
    \item We benchmark with the KITTI dataset and demonstrate competitive results in long-range re-localization.
\end{itemize}
\section{RELATED WORK}

\subsection{Point cloud-based Place Recognition}

Large-scale place recognition is usually formulated as a Bag of Words (BoW) problem, identifying revisits by querying the closest candidate match in a compact latent space. Handcrafted features have shown to be efficient across different domains~\cite{kim2021scan}, but can be sensitive to viewpoint rotation~\cite{Pramatarov2022BoxGraphSP} and translation~\cite{li2021ssc}. 
Conversely, deep learning methods~\cite{komorowski2021minkloc++,vid2022logg3d,komorowski2022egonn,cattaneo2022lcdnet} have demonstrated robustness in retrieving meaningful local and global descriptors simultaneously, but struggle with scalability and domain changes~\cite{keetha2023anyloc}. 

Object-based localization addresses these issues by balancing reduced keypoint extraction with context-informed features~\cite{yuan2024btc,yin2024outram,zhang2023instaloc}. SemSegMap~\cite{Cramariuc2021SemSegMap3} effectively identifies revisits using a CNN to encode each semantic segment while pooling a global descriptor using k-nearest neighbors. Several works improve the sensitivity to occlusion and viewpoint by representing the scene's structure in bird's eye view~\cite{luo2023bevplace} or as edges in a graph~\cite{kong2020sgpr,wang2024sglc}, where each object instance is a node. However, instead of a global embedding per scan, these methods output similarity scores for pairs of scans, making them impractical for real-world robotics that use BoW querying. 

To address the scalability and representation issues, we transform each submap to a graph, where the vertices contain learned rotational-invariant descriptions for each object in the scene. We inform these local embeddings of their neighboring structure using a graph neural network. We also aggregate a global embedding using learned weights that balance the significance of each dimension of the local descriptors.

%To address scalability and representation issues, REGRACE transforms each submap to a graph {\color{red}with learned rotational-invariant descriptions for each object in the scene as vertices.  A graph neural network informs those embeddings of their local neighborhood structure, while a global embedding is computed} using learned weights that balance the significance of each dimension of the local descriptors.

\subsection{Metric Evaluation}

Recent studies~\cite{ kong2020sgpr, wang2024sglc,vidanapathirana2023spectral, hausler2021patch} improve place recognition results by re-ranking the closest matches in the embedding space using contextual geometric cues. Still, they overlook that the short-range physical distance may not align with a linear threshold in the embedding space.  In conventional place recognition evaluation, two candidates represent a loop closure if their embedding distance is below a predefined hard threshold. We address this incompatibility using geometric cues to directly classify revisits. We demonstrate that this evaluation strategy improves performance, particularly in areas that share similar local features but have different structural placements of the objects.

\subsection{Registration}

Given the high number of measurements in a point cloud, performing registration based on all available points is computationally expensive, even more with submap-based representations. To address this issue, several methods propose to perform registration based on a much smaller set of keypoints extracted from these measurements. 
Earlier works such as D3Feat~\cite{bai2020d3feat} select keypoints from a dense cloud based on a saliency score, while recent research leverages sparse point cloud superpoint extraction~\cite{qin2022geometric}. Other approaches~\cite{qin2022geometric, aoki2019pointnetlk} retrieve matching keypoint pairs instead of the individual descriptors, reducing the time spent on feature extraction. Graph-based registration methods use quadratic assignment solvers, such as linear programming or coarse node matching~\cite{ Pramatarov2022BoxGraphSP,wang2024sglc}, but face longer runtimes as node count grows.

RANSAC~\cite{ransac} is a robust realignment method that minimizes the influence of outliers through iterative updates. Although it does not require training data, a more precise estimation can be achieved when combined with ICP~\cite{icp}.
To match objects between two match candidates, we need consistent local features from the same instance, even from different viewpoints. REGRACE extracts rotation-invariant local features appropriate for this realignment.
\section{METHODOLOGY}

\subsection{Problem Formulation}
Assume a set of $K$ submaps $\mathcal{S} = \left\{\mathcal{S}_1, \mathcal{S}_2, \ldots \mathcal{S}_K\right\}$, each represented by a pointcloud $\mathcal{S}_i\in\mathbb{R}^{N_p\times 3}$ in a local map frame $\cframe{M_i}$ and a rigid body transformation $\T{W}{S_i} \in SE(3)$ that transforms points from $\cframe{M_i}$ to a reference world frame $\cframe{W}$, common for all submaps. The re-localization problem consists of querying if the current submap $\mathcal{S}_{c}$  has mapped the same physical region as any of the previous seen submaps in $\mathcal{S}$. If a submap $\mathcal{S}_j \in \mathcal{S}$ represents the same environment as $\mathcal{S}_c$, then a rigid transformation $\T{S_c}{S_j}$ that aligns both submaps is computed.

\subsection{Overview}

\begin{figure*}[t]
      \centering
      \includegraphics[width=\textwidth]{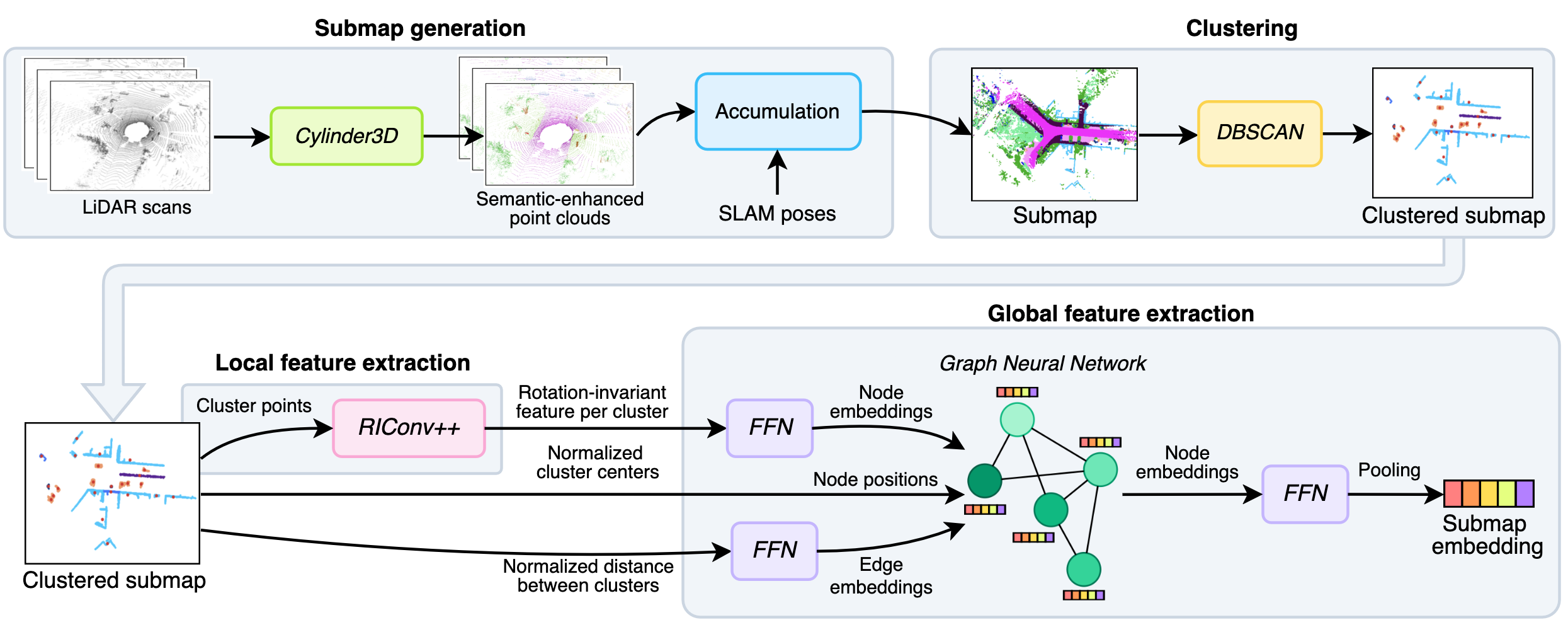}
      \caption{Overview of REGRACE feature extractor. It semantically labels each scan using Cylinder3D~\cite{zhu2020cylindrical} and merges multiple scans into a submap using the poses retrieved by a SLAM pipeline. It then clusters the submap into object instances with DBSCAN~\cite{dbscan}, where each cluster is embedded using RIConv++~\cite{zhang2022riconv2}. Finally, REGRACE polls a unified global descriptor from the node embeddings. Feed-Forward Networks (FFN) connect the encoders.}
      \label{fig:REGRACE} 
\end{figure*}

The REGRACE feature extractor is depicted in Fig.~\ref{fig:REGRACE}. We start by semantically classifying each LiDAR scan and fusing it into a larger submap. Next, we cluster the objects and use convolutional and graph neural networks to extract a local descriptor for each instance. Finally, we create a global descriptor by averaging all object features using a learning-based pooling. For place recognition, we retrieve the top-N closest global descriptors from a database of previously seen submaps and select the most geometrically consistent submap as a revisit candidate. To determine $\boldsymbol{T}_{\mathcal{S}_{\text{c}}\mathcal{S}_j}$ between the query  $\mathcal{S}_{\text{c}}$ and the candidate match $\mathcal{S}_j \in \mathcal{S}$, we register the submaps using the local geometric-informed descriptors.

\subsection{Submap Generation}

Most place recognition research relies on a single sensor, which can be heavily influenced by the viewpoint and data modality. To gather geometric data about object instances and alleviate viewpoint differences, we create dense 3D submaps from a series of LiDAR scans. We use Cylinder3D~\cite{zhu2020cylindrical} to extract semantic labels from each scan. To reduce classification noise, we combine $N$ individual LiDAR scans into a single reference frame using the poses $\boldsymbol{T}_{W\mathcal{S}_i}$ given by a SLAM algorithm.  The number of accumulated scans $N$ depends on the distance between the $N$th scan and the first scan of the submap, which we limit to 20 meters. 

To ensure scalability to large maps, we discretize the submap $\mathcal{S}$ in ten-centimeter voxels. The semantic probability of a voxel is the average of the probabilities of all points inside it, smoothing out erroneous predictions. 

\subsection{Local Features} \label{sec:local-features}

\hyphenation{DBSCAN}

REGRACE subdivides each submap into a set of $K$ point clouds $\mathcal{O} = \left\{\mathcal{O}_1, \mathcal{O}_2, \ldots, \mathcal{O}_K\right\}$, where $\mathcal{O}_i\in\mathbb{R}^{N_i\times 3}$ represents the point cloud of an object in the submap. To cluster voxels with the same semantic label into instances, we use DBSCAN~\cite{dbscan}, a density-based clustering method. {DBSCAN} identifies closely packed voxels separated by lower-density areas using the minimum number of voxels $\eta$ and the maximum distance $\epsilon$ for voxels to belong to the same cluster. This method can detect arbitrarily shaped clusters while eliminating noisy voxels that are not part of any instance. We discard labels that do not constitute closed objects, such as ``road'', ``sidewalk'', ``ground'', ``terrain'', ``unlabeled'', ``outlier'', or ``other-object'' for the semantic classes predicted by Cylinder3D~\cite{zhu2020cylindrical}. These can be extended to other classes based on the application of REGRACE.

We use the center of each object cloud $\boldsymbol{c}_i$ as keypoint coordinates, and learn object features $\boldsymbol{f}_i=\phi_\mathcal{F}(\mathcal{O}_i)$ with $\phi_\mathcal{F}$ being a neural network. In REGRACE, $\phi_\mathcal{F}$ is based on RIConv++~\cite{zhang2022riconv2}, a rotational-invariant adaptation of PointNet++~\cite{qi2017pointnetplusplus}. Our network consists of a sparse convolutional encoder, which outputs object features $\boldsymbol{f} \in \mathbb{R}^{K\times 128}$, where $K$ is the number of objects in the submap. We sample $P=1,024$ farthest points for each cluster. If the point cloud $\mathcal{O}_i$ contains fewer points than $P$, we pad the available $p$ points with the $(P-p)$ farthest points in the cluster.

\subsection{Global descriptor} 
\label{sec:global-features}

Before accumulating the local descriptors into one global feature, we enhance each cluster embedding $\boldsymbol{f}_i$ with context from its local neighborhood using a Graph Neural Network (GNN). This process helps the global feature represent not only the class and amount of individual objects but also how these objects relate to each other in space.  We then pool all node embeddings into a single descriptor per submap. 

\subsubsection{Graph Construction}

We represent each submap by a graph $\mathcal{G}$ containing a set of nodes $\mathcal{V}=\{v_1, v_2, \ldots,v_K\}$ and edges $\mathcal{E}=\{{e}_{12}, {e}_{13}, \ldots,e_{K(K-1)}\}$. Each vertex $v_i=\langle \boldsymbol{c}_i,\boldsymbol{f}_i\rangle$ represents an object, where $\boldsymbol{c}_i$ and $\boldsymbol{f}_i$ are the keypoint coordinates and features defined in Section \ref{sec:local-features}. The edge embeddings $e_{ij}\in \mathbb{R}$ are the normalized Euclidean distance between the corresponding nodes
$$
e_{ij}=\left\|\boldsymbol{c}_i-\boldsymbol{c}_j\right\|_{L2}/\alpha, \eqno{(1)}
$$
where $\alpha$ is the 95\% quantile of the furthest distance between voxels within the same instance in the training split.

While previous works~\cite{kong2020sgpr,wang2024sglc} connect only the k-nearest neighbors (k-NN) to each node, we create a fully connected graph. Therefore, large clusters prevalent at the edges of the submap (i.e., buildings) can transmit their messages to the nodes closer to the center (i.e., cars and poles), resulting in a more distinct and informative global embedding.

\subsubsection{Network Architecture} 
Being robust to viewpoint differences is essential to place recognition. We enhance the local descriptors using an equivariant GNN (EGNN)~\cite{satorras2021n}, which is robust to rotation and translation of the node positions. Our network outputs a graph $\tilde{\mathcal{G}}=\text{EGNN}(\mathcal{G})$ with nodes $\tilde{v}_i=\langle \tilde{\boldsymbol{c}}_i,\tilde{\boldsymbol{f}}_i\rangle\in \tilde{\mathcal{V}}$, where $|\tilde{\mathcal{V}}|=\left|{\mathcal{V}}\right|=K$, $\tilde{\boldsymbol{c}}_i\in \mathbb{R}^{3}$ and ${\tilde{\boldsymbol{f}}}_i\in\mathbb{R}^{512}$. Note that $\tilde{\mathcal{G}}$ and ${\mathcal{G}}$ have the same number of nodes, but with higher-dimensional node embeddings and updated centroid positions. We pool a global descriptor per graph using generalized-mean pooling (GeM)~\cite{radenovic2018fine}
$$
\boldsymbol{g} = \text{GeM}(\tilde{\mathcal{G}}) = \left(\frac{1}{K}\sum_{i=1}^{K}(\tilde{\boldsymbol{f}}i)^{{\lambda}}\right)^{{{\lambda}^{-1}}}, \eqno{(2)}
$$
where ${\lambda}$ is a learnable 1D parameter and $\boldsymbol{g}\in\mathbb{R}^{256}$. 

Like Kong \emph{et al.}~\cite{kong2020sgpr}, we add a second evaluation head to  compute the similarity score $s(\boldsymbol{g}_i,\boldsymbol{g}_j)$ between two submap embeddings using a Tensor Neural Network (TNN)~\cite{socher2013reasoning}
$$
s(\boldsymbol{g}_i,\boldsymbol{g}_j)=\sigma\left(\text{ReLU}\left(\boldsymbol{g}_i^T\boldsymbol{\omega}^{[1:s]}\boldsymbol{g}_j+\boldsymbol{\alpha}\begin{bmatrix}
    \boldsymbol{g}_i\\\boldsymbol{g}_j
\end{bmatrix}+b\right)\right), \eqno{(3)}
$$
where $\boldsymbol{\omega}^{[1:S]}$ and $\boldsymbol{\alpha}$ are learned weights, $b$ a bias factor, $\sigma$ a sigmoid function, and $s=16$ is the size of tensor slices. This head is used only during training.

\subsection{Training Loss}

For place recognition, we use a triplet margin contrastive loss to enforce similarity between the embeddings of revisits
$$
\mathcal{L}_{\text{triplet}} = \max\big(\|\boldsymbol{g}_a-\boldsymbol{g}_p\|_{L2}-\|\boldsymbol{g}_a-\boldsymbol{g}_n\|_{L2}+m\,,\,0\big), \eqno{(4)}
$$
where $\boldsymbol{g}_a$, $\boldsymbol{g}_p$, $\boldsymbol{g}_n$ are the global embeddings of the anchor $\mathcal{S}_a$, positive $\mathcal{S}_p$ and negative  $\mathcal{S}_n$ submaps, respectively. Like in LoGG3D-Net~\cite{vid2022logg3d}, we consider $\mathcal{S}_p$  at most 3m from $\mathcal{S}_a$, whereas $\mathcal{S}_n$ is at least 20m away from $\mathcal{S}_a$.

To enhance performance, we mine the most challenging triplets. The hardest positive and negative examples are the ones furthest and closest to the anchor in the global feature space, respectively. Since this mining requires a forward pass to compute distances, we sample examples within a batch. 

The TNN head is trained using binary cross-entropy loss
$$\mathcal{L}_{\text{score}} = -\big[s_{ij}\log(l_{ij})+(1-s_{ij})\log(1-l_{ij})\big], \eqno{(5)}
$$
where $s_{ij}=s(\boldsymbol{g}_i,\boldsymbol{g}_j)$ is the similarity score between two submaps $\mathcal{S}_i$ and $\mathcal{S}_j$, and $l_{ij}$ the proximity label. If the two submaps are less than 3 meters apart, $l_{ij}=1$; otherwise, $l_{ij}=0$. Our total loss can be expressed as
$$
\mathcal{L} = \mathcal{L}_{\text{triplet}}+\mathcal{L}_{\text{score}}. \eqno{(6)}
$$

\subsection{Loop Closure Detection}

Previous place recognition pipelines classify scan pairs as loop closures if their embedding euclidean distance $D$ is below a predetermined threshold $\delta$. LoGG3D-Net classifies true revisits when the scans are positioned within 3m, while EgoNN~\cite{komorowski2022egonn} uses 5m and 20m. However, the relationship between global embeddings and actual distances is nonlinear, which makes a fixed threshold for identifying loop closure inefficient. To avoid triggering map optimization in areas that are not a loop closure, this threshold $\delta$ is often set lower than the pipeline's capability of identifying hard revisits, increasing the number of false negatives. Moreover, submaps can have similar embeddings while still being farther apart than single LiDAR scans, which leads to false positives. 

To reduce these false detections, we propose using a geometric criterion to classify loop closures. If the geometric consistency score  $C(\mathcal{G}_{i},\mathcal{G}_j)>\varepsilon$, the submap pair $(\mathcal{S}_i, \mathcal{S}_j)$ is considered a revisit. Note that the minimum consistency $\varepsilon$ between two graphs differs from the maximum distance between their global descriptors in the embedding space $\delta$. For each new submap, we re-rank the top 20 closest previous submaps based on embedding similarity $D$, and then select the one with the highest geometric consistency $C$ as the best revisiting candidate. The consistency score  $C(\mathcal{G}_i,\mathcal{G}_j)$~\cite{vidanapathirana2023spectral}  is directly related to the registration success between the query submap $\mathcal{G}_i$ and a submap $\mathcal{G}_j$ in the top 20 closest embeddings as in
$$
C(\mathcal{G}_{i},\mathcal{G}_j) = \sum_{\mathcal{I}} \text{max}\Bigg(
1-\frac{(d_1-d_2)^2}{(d_t)^2},\,0\Bigg), \eqno{(7)}
$$
for $d_1=||\boldsymbol{c}_i-\boldsymbol{c}_j||_{L2}$ and $d_2=||\boldsymbol{c}_{i^\prime}-\boldsymbol{c}_{j^\prime}||_{L2}$, where $(v_i,v_i^\prime)\in\mathcal{G}_{i}$ and $(v_j,v_j^\prime)\in\mathcal{G}_{j}$ are a matching pair of vertices from  RANSAC~\cite{ransac}, $\mathcal{I}$ is the set of inlier pairs of this {RANSAC} estimation, and $d_t$ is the maximum geometric error between the inlier pairs. 
For re-ranking we do not run the dense {ICP} step. We assume $d_t=1$. The RANSAC realignment is further described in the following section.

\subsection{Registration}

REGRACE estimates the transformation between two submaps in a coarse-to-fine approach. To reduce the computation requirements, we initialize the transformation via RANSAC and then proceed to a more demanding ICP~\cite{icp} step once an initial solution is obtained. For the RANSAC step, we realign the object center coordinates $\boldsymbol{c}_i$ rather than the full point cloud. We match keypoints based on the corresponding node features from $\mathcal{G}_i$ and $\mathcal{G}_j$. A match happens if both nodes are mutually the best match between all the nodes of both graphs. Once the transform $\boldsymbol{T}_{\mathcal{S}_\text{c}\mathcal{S}_j}^{\text{RANSAC}}$ has been initialized using RANSAC, we trigger an ICP optimization based on all points in the segmented clusters to retrieve a refined transformation $\boldsymbol{T}_{\mathcal{S}_\text{c}\mathcal{S}_j}^{\text{ICP}}$. To reduce the computational complexity of the ICP step, we register only the objects considered as inliers in the RANSAC step.
\section{EXPERIMENTS}
\subsection{Datasets}
For our benchmark, we use SemanticKITTI~\cite{behley2019semkitti} dataset, which semantically annotates the 360-field-of-view automotive LiDAR scans in the KITTI Odometry dataset~\cite{geiger2012kitti}. KITTI contains 11 sequences of Velodyne HDL-64E scans in urban areas. The ground truth vehicle poses were refined using the SuMa++~\cite{Chen2019SuMaEL} SLAM pipeline. 

Like LoGG3D-Net~\cite{vid2022logg3d}, we train on sequences 00 to 10 in a leave-one-out approach. We evaluate sequences 00, 02, 05, 06, and 08. KITTI-08 contains reverse loop closures. KITTI-07 only contains loop closure between the last and first LiDAR scans. During the submap construction, we take the frame of the middle scan as the origin of the submap. Therefore, the latest submap of KITTI-07 is located 10 meters away from the first submap in the sequence, and there is no valid revisit to report in a submap strategy.

\subsection{Implementation Details}
We trained REGRACE on a NVIDIA A40 GPU with Intel Xeon Gold 6254 @3.1GHz. 
The parameters used for DBSCAN were $\varepsilon=0.05$m and $\eta=800$. For KITTI-02, we used more flexible parameters ($\varepsilon=0.1$m, $\eta=300$) as vegetation covers the background in most of the scans, making it difficult to gather large clusters. In the triplet loss $\mathcal{L_{\text{triplet}}}$, we used the margin $m=1$. We trained using Adam, with $10^{-4}$ learning rate over 100 epochs. On epochs 50 and 75, the learning rate decays by a factor of 10. In a final refinement step, we froze the RIConv++ network and trained only the EGNN over 50 epochs with a learning rate of $10^{-5}$. 

\subsection{Metrics and Evaluation}
ScanContext and many following approaches~\cite{kim2021scan,komorowski2022egonn,kong2020sgpr,wang2024sglc} report evaluation results based on a limited set of candidate pairs. However, in practical applications, a BoW search is used to query a database of previously visited locations. This method allows querying any previously seen submap as a potential candidate for revisit. Like LoGG3D-Net~\cite{vid2022logg3d}, we avoid matching with the same place by excluding adjacent entries less than 30 seconds before the query. 

We evaluate against LoGG3D-Net~\cite{vid2022logg3d} following their 3m and 20m thresholds to classify true and false positives, respectively. We also compare to EgoNN~\cite{komorowski2022egonn}, employing their scheme of 5m and 20m thresholds for true positives on KITTI-00 and 08. The lower the true positive threshold, the closer two submaps must be to be considered a match. By increasing this threshold, we can assess how far in advance the algorithm can identify a potential loop closure. As an object-centric baseline, we compare to SGPR~\cite{kong2020sgpr}. We report Recall@$1$, Recall@$5$, and $F1_{\text{max}}$. We iterate over values of $\delta$ to generate the precision and recall pairs for the $F1_{\text{max}}$.

To evaluate the 6-DoF pose estimation isolated from the place recognition results, we register all submaps within 20 meters of each other, regardless of whether they were classified as revisited by the place recognition pipeline. Similar to EgoNN~\cite{komorowski2022egonn}, we report the mean relative rotation (RRE) and translation (RTE) error for successful registrations, where a successful registration is where RRE $\leq 5.0^\circ$ and RTE $\leq$ 2m. We also report the accuracy score on registration success.

\section{RESULTS}
\subsection{Performance}

Tables~\ref{tab:pr-3m} and~\ref{tab:pr-5-20m} present REGRACE against the selected baselines trained on single scans. 
As EgoNN~\cite{komorowski2022egonn} and SGPR~\cite{kong2020sgpr} output local features, we also report their results using re-ranking of the top 20 closest embeddings and our proposed consistency criterion in Tables~\ref{tab:pr-3m-CE} and~\ref{tab:pr-5-20m-CE}. Since SGPR~\cite{kong2020sgpr} does not output a global embedding, we use the inverse of the similarity score as the embedding distance between two candidate submaps. We also disclose the original results from the SGPR~\cite{kong2020sgpr} paper, which is evaluated on a limited set of submap pairs representing only 30\%  of the total queries following a BoW search method. %We report place recognition results using 3m, 5m and 20m thresholds.

\begin{table}[t]
\begin{center}
\renewcommand{\arraystretch}{1.1}\setlength{\tabcolsep}{6.5pt}
\caption{Place recognition $F1_{\text{max}}$ score with 3m threshold for true positives in  KITTI dataset} \label{tab:pr-3m} 
\begin{tabular}{l|ccccc|c}
\toprule
\multicolumn{1}{c|}{Method} & \multicolumn{1}{c|}{00} & \multicolumn{1}{c|}{02} & \multicolumn{1}{c|}{05} & \multicolumn{1}{c|}{06} & 08 & Avg.\\ \hline
LoGG3D\cite{vid2022logg3d} & \multicolumn{1}{c|}{95.3} & \multicolumn{1}{c|}{{88.8}} & \multicolumn{1}{c|}{97.6} & \multicolumn{1}{c|}{97.7} & 84.3 & 92.7\\
SGPR$^\star$\cite{kong2020sgpr} & \multicolumn{1}{c|}{96.9} & \multicolumn{1}{c|}{89.1} & \multicolumn{1}{c|}{90.5} & \multicolumn{1}{c|}{97.1} &  90.0 & 92.7\\ 
SGPR\cite{kong2020sgpr} & \multicolumn{1}{c|}{8.8} & \multicolumn{1}{c|}{23.3} & \multicolumn{1}{c|}{58.6} & \multicolumn{1}{c|}{63.7} &   0.1 & 30.9\\ 
EgoNN\cite{komorowski2022egonn}  & \multicolumn{1}{c|}{98.1} & \multicolumn{1}{c|}{\textbf{89.8}} & \multicolumn{1}{c|}{\textbf{97.8}} & \multicolumn{1}{c|}{{99.8}} & \textbf{85.1} & \textbf{94.1}  \\ 
\rowcolor[HTML]{cffafe} REGRACE (ours) & \multicolumn{1}{c|}{\textbf{99.2}} & \multicolumn{1}{c|}{88.9} & \multicolumn{1}{c|}{\textbf{97.8}} & \multicolumn{1}{c|}{\textbf{100.0}} & 82.6 &{93.7} \\ \hline
\end{tabular}
\end{center} 
{\footnotesize 
\hphantom{x}$\star$ original paper results} %
\end{table}

\begin{table}[t]
\begin{center}
\renewcommand{\arraystretch}{1.1}\setlength{\tabcolsep}{5.5pt}
\caption{Place recognition $F1_{\text{max}}$ with 3m for true positives, re-ranking, and our consistency criterion in KITTI dataset} \label{tab:pr-3m-CE} 
\begin{tabular}{l|ccccc|c}
\toprule
\multicolumn{1}{c|}{Method} & \multicolumn{1}{c|}{00} & \multicolumn{1}{c|}{02} & \multicolumn{1}{c|}{05} & \multicolumn{1}{c|}{06} & 08 & Avg.\\ \hline
SGPR\cite{kong2020sgpr} & \multicolumn{1}{c|}{8.8} & \multicolumn{1}{c|}{23.3} & \multicolumn{1}{c|}{58.6} & \multicolumn{1}{c|}{63.7} &   0.1 & 30.9\\ 
SGPR\cite{kong2020sgpr} + RR & \multicolumn{1}{c|}{\underline{18.6}} & \multicolumn{1}{c|}{\textbf{51.9}} & \multicolumn{1}{c|}{\underline{74.4}} & \multicolumn{1}{c|}{\underline{93.2}} &  \textbf{2.9} & \underline{48.2}\\  
\rowcolor[HTML]{cffafe} SGPR\cite{kong2020sgpr} + RR/CE& \multicolumn{1}{c|}{\textbf{61.1}} & \multicolumn{1}{c|}{\underline{49.7}} & \multicolumn{1}{c|}{\textbf{79.6}} & \multicolumn{1}{c|}{\textbf{94.4}} & \underline{0.2} & \textbf{57.0}\\ 
\hline
EgoNN\cite{komorowski2022egonn}  & \multicolumn{1}{c|}{\underline{98.1}} & \multicolumn{1}{c|}{\underline{89.8}} & \multicolumn{1}{c|}{\underline{97.8}} & \multicolumn{1}{c|}{\underline{99.8}} & \underline{85.1} & \underline{94.1}  \\
EgoNN\cite{komorowski2022egonn} + RR& \multicolumn{1}{c|}{97.5} & \multicolumn{1}{c|}{{87.7}} & \multicolumn{1}{c|}{95.8} & \multicolumn{1}{c|}{\underline{99.8}} & 82.1 & 92.6 \\ 
 \rowcolor[HTML]{cffafe} EgoNN\cite{komorowski2022egonn} + RR/CE& \multicolumn{1}{c|}{\textbf{100.0}} & \multicolumn{1}{c|}{\textbf{94.3}} & \multicolumn{1}{c|}{\textbf{99.5}} & \multicolumn{1}{c|}{\textbf{100.0}} & \textbf{89.2} &\textbf{96.6} \\
\hline
% REGRACE$^\ast$& \multicolumn{1}{c|}{80.6} & \multicolumn{1}{c|}{79.8} & \multicolumn{1}{c|}{84.1} & \multicolumn{1}{c|}{76.9} & 22.9 &68.7 \\
REGRACE$^\dagger$ & \multicolumn{1}{c|}{84.6} & \multicolumn{1}{c|}{82.3} & \multicolumn{1}{c|}{83.8} & \multicolumn{1}{c|}{\textbf{100.0}} & 34.7 & 77.1 \\ 
\rowcolor[HTML]{cffafe} REGRACE (ours) & \multicolumn{1}{c|}{\textbf{99.2}} & \multicolumn{1}{c|}{\textbf{88.9}} & \multicolumn{1}{c|}{\textbf{97.8}} & \multicolumn{1}{c|}{\textbf{100.0}} & {\textbf{82.6}} &{\textbf{93.7}} \\ \hline
\end{tabular} 
\end{center}
{\footnotesize 
%\hphantom{x}$\ast$ without re-ranking or consistency evaluation\\
\hphantom{x}$\dagger$ REGRACE with re-ranking but without consistency evaluation}
\end{table}

\begin{figure}[t]
      \centering
      \includegraphics[width=\linewidth]{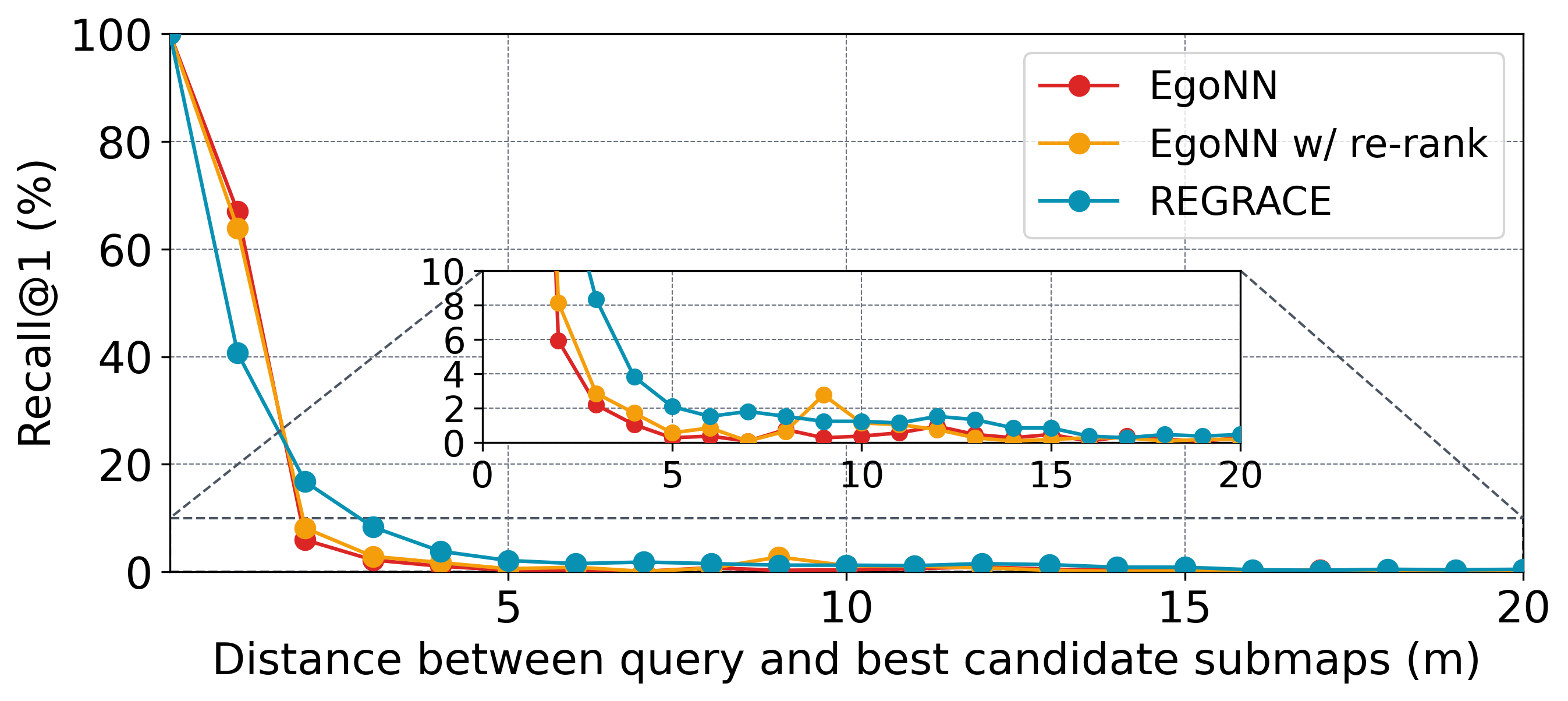}
    \caption{Recall@1 between query and best candidate submap pairs using BoW search in KITTI-00 for the 20m threshold. } \label{fig:loc-results-recall} 
\end{figure}

\begin{figure}[t]
      \centering
        \begin{minipage}{0.24\textwidth}
            \centering
            \includegraphics[width=\textwidth]{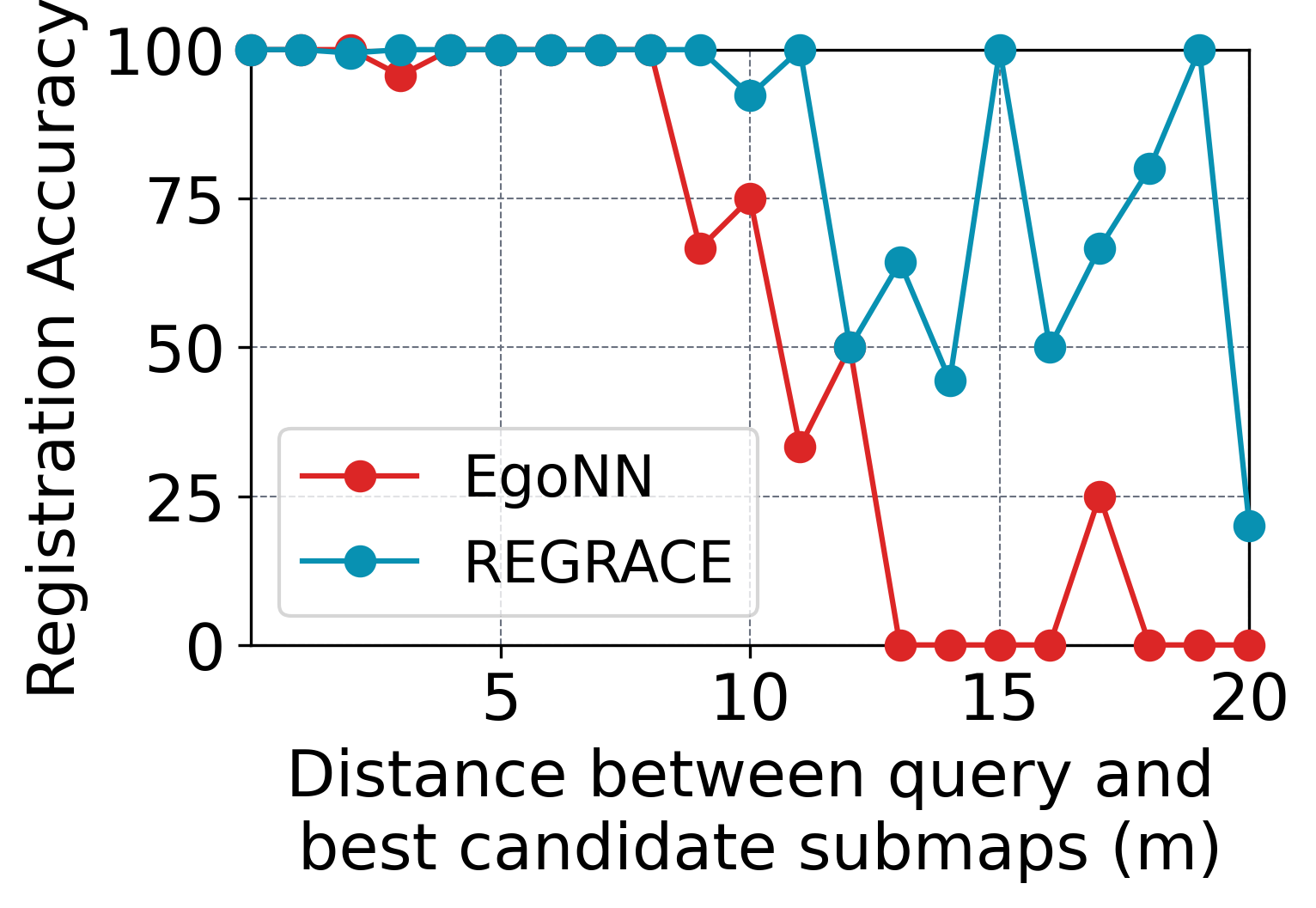}
        \end{minipage}\hfill
        \begin{minipage}{0.24\textwidth}
            \centering
            \includegraphics[width=\textwidth]{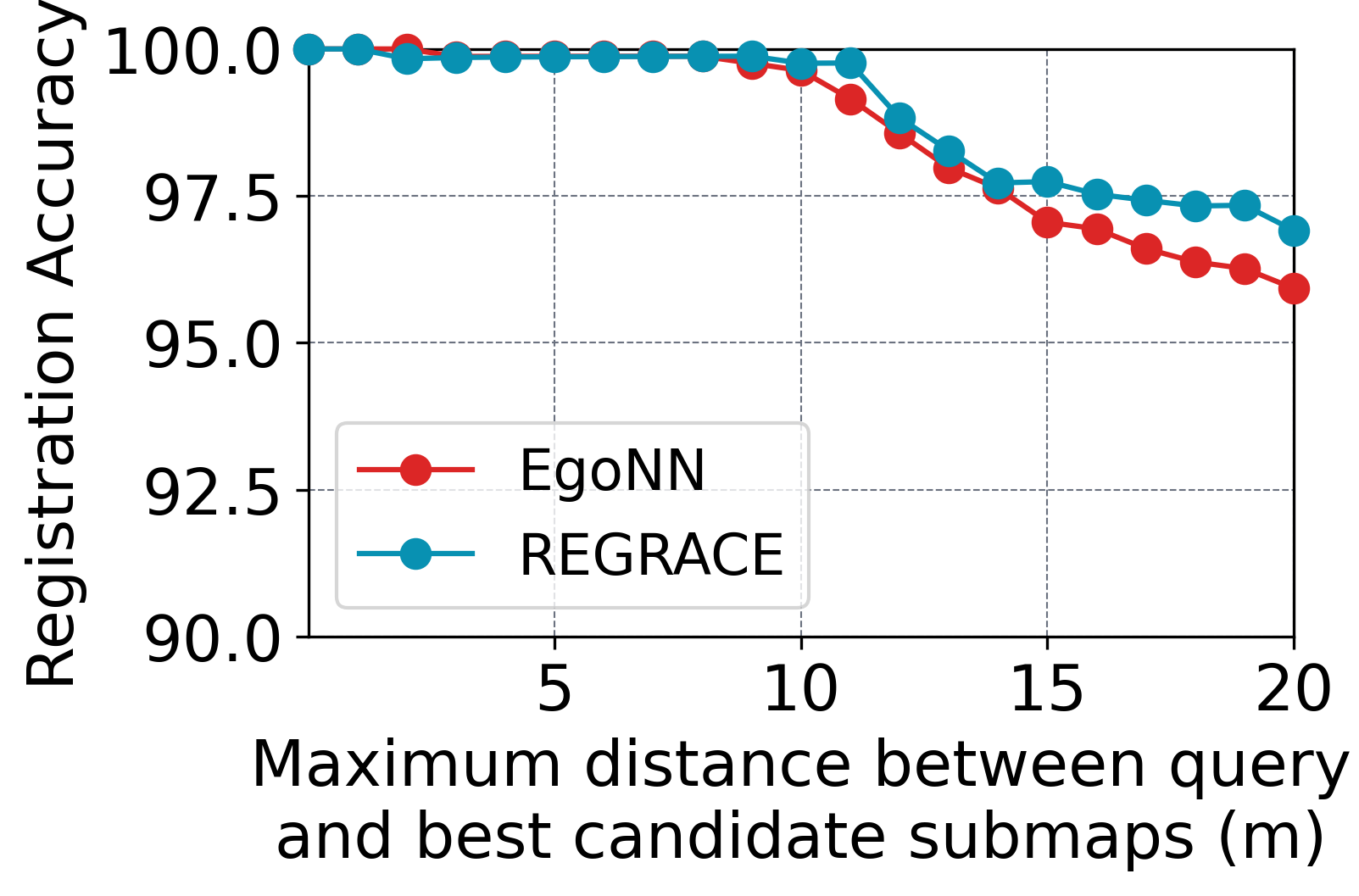}
        \end{minipage}
        \caption{Registration accuracy versus distance between the query and best candidate submap pairs using BoW search in KITTI-00. The cumulative sum of (left) is depicted in (right).} 
        \label{fig:loc-results-reg} 
\end{figure}

\begin{figure}[t]
      \centering
      \includegraphics[width=\linewidth]{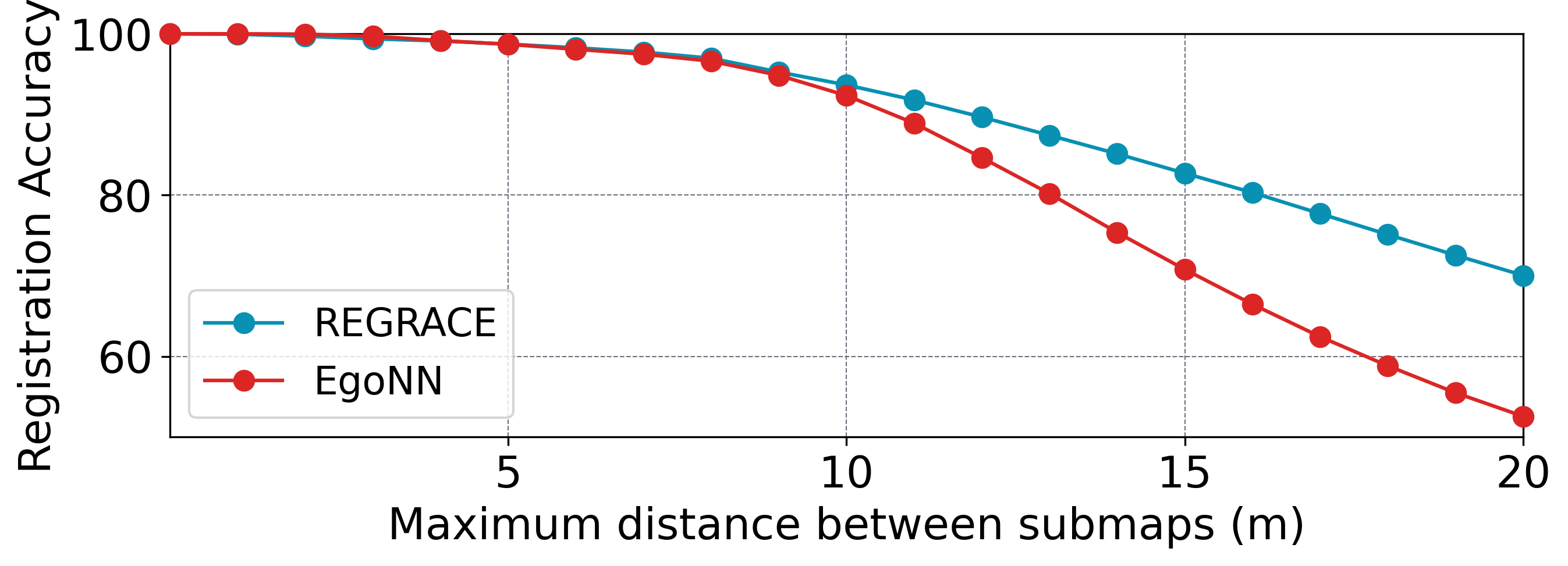}
    \caption{Registration accuracy versus distance between query and candidate submap pairs for the registration test set in KITTI-00. } \label{fig:reg-results}
\end{figure}

\begin{table*}[t]
\renewcommand{\arraystretch}{1.1}
\begin{center}
\setlength{\tabcolsep}{8pt}
\caption{Place recognition recall and $F1_{\text{max}}$ scores with 5m and 20m threshold for true positives in  KITTI}
\label{tab:pr-5-20m}
\begin{tabular}{l|cccccccccccc}
\toprule
\multicolumn{1}{c|}{\multirow{3}{*}{Method}} & \multicolumn{6}{c|}{00} & \multicolumn{6}{c}{08} \\ 
 & \multicolumn{3}{c|}{5m} & \multicolumn{3}{c|}{20m} & \multicolumn{3}{c|}{5m} & \multicolumn{3}{c}{20m} \\ \cline{2-13} 
 & \multicolumn{1}{c|}{R@1} & \multicolumn{1}{c|}{R@5 } & \multicolumn{1}{c|}{$F1_{\text{max}}$} & \multicolumn{1}{c|}{R@1} & \multicolumn{1}{c|}{R@5 } & \multicolumn{1}{c|}{$F1_{\text{max}}$} & \multicolumn{1}{c|}{R@1} & \multicolumn{1}{c|}{R@5 } & \multicolumn{1}{c|}{$F1_{\text{max}}$} & \multicolumn{1}{c|}{R@1} & \multicolumn{1}{c|}{R@5 } & \multicolumn{1}{c}{$F1_{\text{max}}$}  \\  \hline
SGPR\cite{kong2020sgpr} & \multicolumn{1}{c|}{{53.9}} & \multicolumn{1}{c|}{{77.6}} & \multicolumn{1}{c|}{11.1} & \multicolumn{1}{c|}{50.3} & \multicolumn{1}{c|}{73.6}  & \multicolumn{1}{c|}{19.0} & \multicolumn{1}{c|}{3.3} & \multicolumn{1}{c|}{13.9}  & \multicolumn{1}{c|}{1.9} & \multicolumn{1}{c|}{39.1} & \multicolumn{1}{c|}{69.7}  & \multicolumn{1}{c}{22.2}\\ 
EgoNN\cite{komorowski2022egonn} & \multicolumn{1}{c|}{\textbf{98.0}} & \multicolumn{1}{c|}{\textbf{98.5}} & \multicolumn{1}{c|}{96.5} & \multicolumn{1}{c|}{82.2} & \multicolumn{1}{c|}{86.3}  & \multicolumn{1}{c|}{85.1} & \multicolumn{1}{c|}{\textbf{90.5}} & \multicolumn{1}{c|}{\textbf{97.4}}  & \multicolumn{1}{c|}{83.6} & \multicolumn{1}{c|}{69.0} & \multicolumn{1}{c|}{78.9}  & \multicolumn{1}{c}{69.9} \\
\rowcolor[HTML]{cffafe} REGRACE (ours) & \multicolumn{1}{c|}{92.0} & \multicolumn{1}{c|}{93.5}  & \multicolumn{1}{c|}{\textbf{99.4}} & \multicolumn{1}{c|}{\textbf{87.2}} & \multicolumn{1}{c|}{\textbf{88.4}}  & \multicolumn{1}{c|}{\textbf{90.8}} & \multicolumn{1}{c|}{73.5} & \multicolumn{1}{c|}{84.7}  & \multicolumn{1}{c|}{\textbf{86.7}} & \multicolumn{1}{c|}{\textbf{89.6}} &  \multicolumn{1}{c|}{\textbf{91.9}} & \multicolumn{1}{c}{\textbf{85.2}} \\ \hline
\end{tabular}
\end{center}
\end{table*}

\begin{table*}[t]
\renewcommand{\arraystretch}{1.1}
\setlength{\tabcolsep}{8pt}
\begin{center}
\caption{Place recognition recall and $F1_{\text{max}}$ scores with 5m and 20m thresholds for true positives, top-20 re-ranking, and our proposed consistency criterion in  KITTI dataset}
\label{tab:pr-5-20m-CE}
\begin{tabular}{l|cccccccccccc}
\toprule
\multicolumn{1}{c|}{\multirow{3}{*}{Method}} & \multicolumn{6}{c|}{00} & \multicolumn{6}{c}{08} \\ 
 & \multicolumn{3}{c|}{5m} & \multicolumn{3}{c|}{20m} & \multicolumn{3}{c|}{5m} & \multicolumn{3}{c}{20m} \\ \cline{2-13} 
 & \multicolumn{1}{c|}{R@1} & \multicolumn{1}{c|}{R@5 } & \multicolumn{1}{c|}{$F1_{\text{max}}$} & \multicolumn{1}{c|}{R@1} & \multicolumn{1}{c|}{R@5 } & \multicolumn{1}{c|}{$F1_{\text{max}}$} & \multicolumn{1}{c|}{R@1} & \multicolumn{1}{c|}{R@5 } & \multicolumn{1}{c|}{$F1_{\text{max}}$} & \multicolumn{1}{c|}{R@1} & \multicolumn{1}{c|}{R@5 } & \multicolumn{1}{c}{$F1_{\text{max}}$}  \\  \hline
 
SGPR\cite{kong2020sgpr} & \multicolumn{1}{c|}{{53.9}} & \multicolumn{1}{c|}{{77.6}} & \multicolumn{1}{c|}{11.1} & \multicolumn{1}{c|}{50.3} & \multicolumn{1}{c|}{73.6}  & \multicolumn{1}{c|}{19.0} & \multicolumn{1}{c|}{3.3} & \multicolumn{1}{c|}{13.9}  & \multicolumn{1}{c|}{1.9} & \multicolumn{1}{c|}{\textbf{39.1}} & \multicolumn{1}{c|}{69.7}  & \multicolumn{1}{c}{22.2}\\ 
 SGPR\cite{kong2020sgpr} + RR & \multicolumn{1}{c|}{\textbf{65.7}} & \multicolumn{1}{c|}{\underline{77.7}} & \multicolumn{1}{c|}{\underline{24.5}} & \multicolumn{1}{c|}{\textbf{63.5}} & \multicolumn{1}{c|}{\underline{74.6}}  & \multicolumn{1}{c|}{\underline{36.2}} & \multicolumn{1}{c|}{\textbf{6.1}} & \multicolumn{1}{c|}{\underline{20.6}}  & \multicolumn{1}{c|}{\underline{1.5}} & \multicolumn{1}{c|}{38.4} & \multicolumn{1}{c|}{\underline{72.3}}  & \multicolumn{1}{c}{\underline{36.7}}\\
\rowcolor[HTML]{cffafe} SGPR\cite{kong2020sgpr} + RR/CE & \multicolumn{1}{c|}{\underline{60.8}} & \multicolumn{1}{c|}{\textbf{78.3}} & \multicolumn{1}{c|}{\textbf{65.7}} & \multicolumn{1}{c|}{\underline{63.4}} & \multicolumn{1}{c|}{\textbf{75.3}}  & \multicolumn{1}{c|}{\textbf{43.8}} & \multicolumn{1}{c|}{\underline{5.8}} & \multicolumn{1}{c|}{\textbf{19.5}}  & \multicolumn{1}{c|}{\textbf{2.1}} & \multicolumn{1}{c|}{\underline{38.5}} & \multicolumn{1}{c|}{\textbf{72.4}}  & \multicolumn{1}{c}{\textbf{39.8}}\\  \hline
EgoNN\cite{komorowski2022egonn} & \multicolumn{1}{c|}{{98.0}} & \multicolumn{1}{c|}{{98.5}} & \multicolumn{1}{c|}{\underline{96.5}} & \multicolumn{1}{c|}{82.2} & \multicolumn{1}{c|}{86.3}  & \multicolumn{1}{c|}{\underline{85.1}} & \multicolumn{1}{c|}{{90.5}} & \multicolumn{1}{c|}{{97.4}}  & \multicolumn{1}{c|}{\underline{83.6}} & \multicolumn{1}{c|}{69.0} & \multicolumn{1}{c|}{78.9}  & \multicolumn{1}{c}{\underline{69.9}} \\ 
EgoNN\cite{komorowski2022egonn} + RR& \multicolumn{1}{c|}{\underline{98.8}} & \multicolumn{1}{c|}{\underline{99.3}}  & \multicolumn{1}{c|}{96.1}& \multicolumn{1}{c|}{\textbf{86.5}} & \multicolumn{1}{c|}{\textbf{88.3}}  & \multicolumn{1}{c|}{84.4} & \multicolumn{1}{c|}{\underline{96.6}} & \multicolumn{1}{c|}{\underline{99.1}}  & \multicolumn{1}{c|}{81.5} & \multicolumn{1}{c|}{\textbf{83.2}} & \multicolumn{1}{c|}{\textbf{85.0}}   & \multicolumn{1}{c}{69.1}\\ 
\rowcolor[HTML]{cffafe} EgoNN\cite{komorowski2022egonn} + RR/CE& \multicolumn{1}{c|}{\textbf{98.9}} & \multicolumn{1}{c|}{\textbf{99.4}}  & \multicolumn{1}{c|}{\textbf{99.7}} & \multicolumn{1}{c|}{\underline{86.3}} & \multicolumn{1}{c|}{\textbf{88.3}}  & \multicolumn{1}{c|}{\textbf{87.7}} & \multicolumn{1}{c|}{\textbf{96.9}} & \multicolumn{1}{c|}{\textbf{99.2}}  & \multicolumn{1}{c|}{\textbf{89.9}} & \multicolumn{1}{c|}{\underline{82.9}} & \multicolumn{1}{c|}{\underline{84.6}} & \multicolumn{1}{c}{\textbf{80.7}}\\  \hline
REGRACE$^\dagger$ & \multicolumn{1}{c|}{91.9} & \multicolumn{1}{c|}{93.4}  & \multicolumn{1}{c|}{83.2} & \multicolumn{1}{c|}{\textbf{87.2}} & \multicolumn{1}{c|}{{88.2}}  & \multicolumn{1}{c|}{73.2} & \multicolumn{1}{c|}{71.6} & \multicolumn{1}{c|}{84.4}  & \multicolumn{1}{c|}{42.1} & \multicolumn{1}{c|}{{89.1}} &  \multicolumn{1}{c|}{{91.7}}  & \multicolumn{1}{c}{44.1}\\  
\rowcolor[HTML]{cffafe} REGRACE (ours) & \multicolumn{1}{c|}{\textbf{92.0}} & \multicolumn{1}{c|}{\textbf{93.5}}  & \multicolumn{1}{c|}{\textbf{99.4}} & \multicolumn{1}{c|}{\textbf{87.2}} & \multicolumn{1}{c|}{\textbf{88.4}}  & \multicolumn{1}{c|}{\textbf{90.8}} & \multicolumn{1}{c|}{\textbf{73.5}} & \multicolumn{1}{c|}{\textbf{84.7}}  & \multicolumn{1}{c|}{\textbf{86.7}} & \multicolumn{1}{c|}{\textbf{89.6}} &  \multicolumn{1}{c|}{\textbf{91.9}} & \multicolumn{1}{c}{\textbf{85.2}} \\ \hline
\end{tabular}
\end{center}
{\footnotesize 
\hphantom{xx}$\dagger$ REGRACE with re-ranking and without consistency evaluation}
\end{table*}

REGRACE outperforms LoGG3D-Net~\cite{vid2022logg3d} and SGPR~\cite{kong2020sgpr} in the 3m range, while EgoNN~\cite{komorowski2022egonn} has the highest $F1_{\text{max}}$ in average. Our consistency criterion improves all tested methods, boosting $F1_{\text{max}}$ scores for EgoNN~\cite{komorowski2022egonn} and REGRACE by 4\% and 25\%, respectively, surpassing re-ranking alone. Note that EgoNN's~\cite{komorowski2022egonn} retrieval of 128 keypoints per LiDAR scan gives an advantage in representation power. In contrast, REGRACE segments an average of 20 objects per submap. Submaps enhance our performance at 20 meters, exceeding EgoNN~\cite{komorowski2022egonn} in all metrics. Fig.~\ref{fig:loc-results-recall} and Fig.~\ref{fig:loc-results-reg} show that REGRACE is more robust in a wider range and that our re-ranking strategy enhances the baseline scores.

SGPR~\cite{kong2020sgpr} was trained with binary-cross entropy and does not enforce similarity in the global embedding space. Therefore, our BoW evaluation on SGPR~\cite{kong2020sgpr} scores lower than the original paper results. Moreover, SGPR~\cite{kong2020sgpr} local features, based on bounding boxes, are ambiguous and less effective for registration, failing to distinguish positive examples by consistency, particularly on reverse loop closures in KITTI-08. Our embeddings are distinct enough to achieve 90\% Recall@1 in sequences that present reverse loop closures without removing dynamic objects like in SGPR~\cite{kong2020sgpr}. 

Note that our consistency evaluation (RR + CE) improves $F1_{\text{max}}$ performance of all baselines. For EgoNN~\cite{komorowski2022egonn}, using only re-ranking (RR) reduces the $F1_{\text{max}}$ by 2\% in relation to no re-ranking. Since the new top-1 choice after re-ranking is not guaranteed to have a closer embedding to the query than the original best revisiting candidate, false positives may increase. In contrast, detecting loop closures based on geometric cues guarantees that the new top-1 choice has the best fit to the loop detection criterion than the other top-20 closest embeddings, increasing the overall metrics.

\begin{table*}[t]
\begin{minipage}{.7\linewidth}
\begin{center}
\caption{6-DoF pose estimation results within 2m and 5$^\circ$ threshold in  KITTI dataset}\label{ref:reg}
\renewcommand{\arraystretch}{1.1}
\begin{tabular}{l|cccccc}
\toprule
\multicolumn{1}{c|}{\multirow{2}{*}{Method}} & \multicolumn{3}{c|}{00} & \multicolumn{3}{c}{08}  \\
 & \multicolumn{1}{c|}{Acc(\%) $\uparrow$} & \multicolumn{1}{c|}{RTE(cm) $\downarrow$} & \multicolumn{1}{c|}{RRE($^{\circ}$) $\downarrow$} & \multicolumn{1}{c|}{Acc(\%) $\uparrow$} & \multicolumn{1}{c|}{RTE(cm) $\downarrow$} & RRE($^{\circ}$)  $\downarrow$ \\ \hline
EgoNN\cite{komorowski2022egonn} & \multicolumn{1}{c|}{52.3} & \multicolumn{1}{c|}{16.5} & \multicolumn{1}{c|}{1.23} & \multicolumn{1}{c|}{\textbf{72.4}} & \multicolumn{1}{c|}{24.2} & 2.13  \\ 
\rowcolor[HTML]{cffafe}REGRACE (ours) & \multicolumn{1}{c|}{\textbf{70.1}} & \multicolumn{1}{c|}{\textbf{5.33}} & \multicolumn{1}{c|}{\textbf{0.21}} & \multicolumn{1}{c|}{{69.7}} & \multicolumn{1}{c|}{\textbf{9.04}} & \textbf{0.48} \\ \hline
\end{tabular}
\end{center}
\end{minipage}\hfill
\begin{minipage}{.28\linewidth}
\begin{center}
\caption{Timing analysis}\label{tab:timing}
\renewcommand{\arraystretch}{1.1}\setlength{\tabcolsep}{3.5pt}
\begin{tabular}{l|>{\columncolor[HTML]{cffafe}}c|c}
\toprule
\multicolumn{1}{c|}{{Step}}   & REGRACE & EgoNN~\cite{komorowski2022egonn}\\ \hline
Inference             & \textbf{24.3ms} & 32.8ms   \\ \hline
Re-ranking                  & \textbf{56.2ms} & 146.2ms   \\ \hline
Registration                         & \textbf{56.1ms} & 150.6ms   \\ \hline
\end{tabular}
\end{center}
\end{minipage}
\end{table*}

In Table \ref{ref:reg}, we compare our registration performance using the local features produced by REGRACE and EgoNN~\cite{komorowski2022egonn}. We increase the registration accuracy over 17.7\% in KITTI-00, and have comparable results to EgoNN in KITTI-08.  Fig.~\ref{fig:reg-results} shows that REGRACE effectively registers difficult-to-match distant pairs, consistently identifying distinct geometric features in the embedding space. REGRACE achieves an average registration error of 7.2 cm and 0.35°, 13.2 cm and 1.34° better than EgoNN's~\cite{komorowski2022egonn}.

Table \ref{tab:timing} presents the time, measured in milliseconds, required for re-localization. Although REGRACE uses submaps as inputs instead of raw scans, its speed is twice that of EgoNN~\cite{komorowski2022egonn}. This result demonstrates the scalability of our method in generating fewer but informative embeddings. Overall, REGRACE operates at a frequency of 7.5 Hz.

\subsection{Ablation Studies}

We conducted a series of ablation experiments on KITTI-00 to evaluate the performance of the REGRACE components. The results are summarized in Table \ref{tab:ablations}. We report the metrics for the non-refined network trained over 100 epochs. We classify revisits using embedding distance to isolate the improvements provided by the consistency criterion. Comparing [A] and [B], it is clear that incorporating edges between all nodes produces more informative global submap embeddings. The comparison between [B] and [C] demonstrates that including binary cross-entropy alongside the triplet loss improves submap descriptions, enforcing similarity in the embedding space for corresponding places.

To evaluate the chosen pooling algorithm, we also tested VLAD~\cite{Arandjelovic2018VLAD} pooling using a PointNetVLAD ~\cite{uy2018pointnetvlad} layer and an arithmetic mean.  GeM~\cite{radenovic2018fine} [F] outperforms VLAD ~\cite{Arandjelovic2018VLAD} [E] as the latter encodes the local feature residuals into clusters without accounting for the high variability of the node descriptors. The results from [D] and [F] also show the benefit of learnable weights from GeM~\cite{radenovic2018fine} in controlling the sensitivity of outliers compared to a simple mean across all dimensions of the node embeddings.

\begin{table}[t]
\begin{center}
\renewcommand{\arraystretch}{1.1}
\setlength{\tabcolsep}{3.3pt}
\caption{Ablation studies}\label{tab:ablations}
\begin{tabular}{c|c|c|c}
\toprule
& \multicolumn{1}{c|}{K-NN / Loss} & $F1_{\text{max}}$  & R@1  \\ \hline
{\footnotesize{[A]}} & 10\hphantom( + $\mathcal{L}_{\text{triplet}}$& 66.7 & 71.2  \\ 
{\footnotesize{[B]}} & All + $\mathcal{L}_{\text{triplet}}$ & 71.4 & 71.9  \\ 
 \footnotesize{[C]} & All + $\mathcal{L}$ (6)\hphantom( & \textbf{77.0} & \textbf{74.4} \\ \hline
\end{tabular}\hfill
\begin{tabular}{c|c|c|c}
\toprule
& \multicolumn{1}{c|}{Pooling} & $F1_{\text{max}}$ & R@1 \\ \hline
{\footnotesize{[D]}} & VLAD \cite{Arandjelovic2018VLAD} & 57.1 & 52.3  \\ 
{\footnotesize{[E]}} & Global mean & 59.2 & 55.6  \\ 
{\footnotesize{[F]}} & GeM \cite{radenovic2018fine} & \textbf{71.4} & \textbf{71.9}  \\\hline
\end{tabular} 
\end{center}
\end{table}

\section{CONCLUSIONS}

This work introduces REGRACE, a scalable place recognition and registration pipeline for dense 3D input. REGRACE efficiently recognizes revisits and aligns dense submaps while maintaining scalability. It employs a novel loop closure detection method that emphasizes geometric consistency rather than embedding distance, leading to improved accuracy even on compared baselines. Future research could investigate using attention pooling to create more distinctive global embeddings. Additionally, one could incorporate local consistency loss to enhance the similarity among the embeddings of the same object instance across different submaps.  Applying REGRACE in scene-graph SLAM pipelines could also provide further insights into the graph-representation scalability in long-range mapping.

%%%%%%%%%%%%%%%%%%%%%%%%%%%%%%%%%%%%%%%%%%%%%%%%%%%%%%%%%%%%%%%%%%%%%%%%%%%%%%%%

%%%%%%%%%%%%%%%%%%%%%%%%%%%%%%%%%%%%%%%%%%%%%%%%%%%%%%%%%%%%%%%%%%%%%%%%%%%%%%%%

%%%%%%%%%%%%%%%%%%%%%%%%%%%%%%%%%%%%%%%%%%%%%%%%%%%%%%%%%%%%%%%%%%%%%%%%%%%%%%%%

%%%%%%%%%%%%%%%%%%%%%%%%%%%%%%%%%%%%%%%%%%%%%%%%%%%%%%%%%%%%%%%%%%%%%%%%%%%%%%%%

%%% bibliography
\bibliography{mybib}

\end{document}